\documentclass{svproc}
\usepackage{url}
\usepackage{graphicx}
\usepackage{amsmath}
\usepackage{amssymb}
\usepackage{bm}
\usepackage{booktabs}
\usepackage{algorithm}
\usepackage{algorithmic}

\begin{document}
\mainmatter              
\title{Learning-Accelerated Optimization-based Trajectory Planning for Cooperative Aerial-Ground Handover Missions}
\titlerunning{Learning-Accelerated Trajectory Planning}  
%
\author{Jingshan Chen\inst{1} \and Bochen Yu\inst{1}
Henrik Ebel\inst{2} \and Peter Eberhard\inst{1}}
\authorrunning{Jingshan Chen et al.} 
%
\tocauthor{Jingshan Chen, Bochen Yu, Henrik Ebel, and Peter Eberhard}
\institute{Institute of Engineering and Computational Mechanics, University of Stuttgart, 70569 Stuttgart, Germany\\
\email{jingshan.chen@itm.uni-stuttgart.de and peter.eberhard@itm.uni-stuttgart.de}\\ 
\and
Mechanical Engineering, LUT University, 53850 Lappeenranta, Finland\\
\email{henrik.ebel@lut.fi}
}

\maketitle              

\begin{abstract}
This paper presents a learning-augmented trajectory planning framework for cooperative unmanned aerial vehicle (UAV) and unmanned ground vehicle (UGV) handover missions. While centralized trajectory optimization ensures dynamic feasibility and task optimality, its high computational cost limits real-time applicability. We propose a neural surrogate planner utilizing decoupled encoder-decoder long short-term memory (LSTM) networks to generate coordinated handover trajectory predictions from the task specifications. These predictions serve as informed warm starts for the downstream centralized optimizer, thereby accelerating convergence to dynamically feasible solutions. Benchmark evaluations demonstrate that the learning-augmented planning framework achieves more than a threefold speedup and 100\% optimization success rate compared to cold start optimization. The results indicate that combining data-driven inference with model-based refinement enables fast and reliable trajectory generation for heterogeneous multi-robot systems.

\keywords{cooperative trajectory planning, heterogeneous multi-robot system, warm-started optimization, LSTM networks}
\end{abstract}

\section{Introduction}
Planning kinematically and dynamically feasible trajectories for multiple robots in cooperative missions, while simultaneously optimizing task-related objectives such as completion time and control effort, remains highly challenging. A widely adopted paradigm is to first compute collision-free paths or waypoints at a geometric or symbolic level, and then generate time-parameterized trajectories that interpolate these waypoints~\cite{Siciliano2009,Li2019}. However, in such hierarchical schemes, system dynamics and constraints are often not enforced from the outset, so the resulting trajectories may not be systematically guaranteed to be dynamically feasible, which can complicate subsequent tracking by low-level controllers~\cite{Pham2018}. 
Moreover, trajectories obtained from hand-crafted or heuristically designed paths are rarely optimal for the overall mission, often resulting in conservative solutions that do not fully utilize the system’s capabilities and reduce overall task efficiency~\cite{Madridano2021,Cao2022}.

Our previous work addresses this challenge by formulating a centralized trajectory optimization problem that explicitly incorporates the full system dynamics together with a mathematical description of cooperative tasks to realize an object handover between a UAV and a UGV in a transport scenario~\cite{Chen2024}. In that setting, a ground robot first carries a parcel and a quadrotor must autonomously rendezvous with it, pick up the parcel, and deliver it to a target location that is inaccessible to the ground robot, while the handover timing and location are jointly optimized within a single optimal control problem. Although this optimization-based planner yields dynamically feasible and task-optimal cooperative trajectories, its runtime becomes a critical bottleneck, particularly in dynamic environments that demand frequent trajectory replanning.

\begin{figure}[htb]
  \centering
    \includegraphics[width=\textwidth]{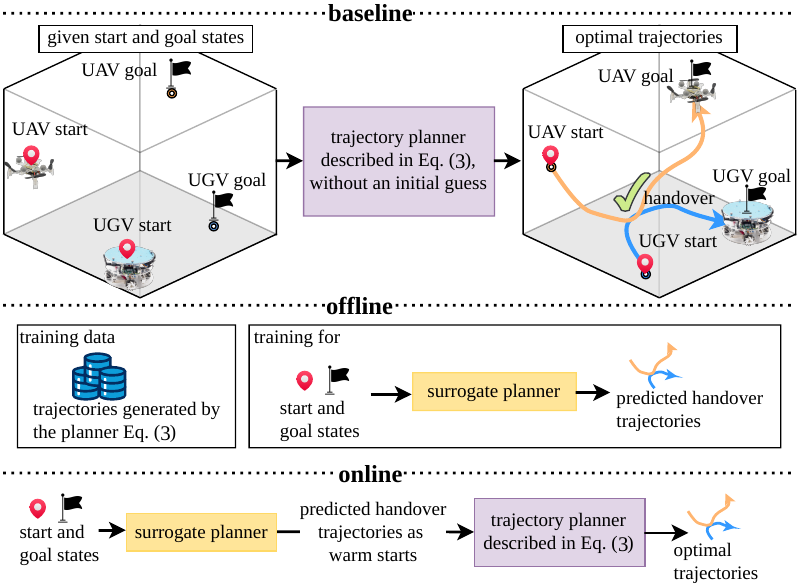}
    \caption{Overview of the proposed learning-accelerated trajectory planning pipeline. The baseline planner generates expert demonstrations for surrogate training offline, and the trained surrogate then provides informed warm starts to the same planner during online deployment.}
  \label{fig:learning_augmented_overview}
\end{figure}

To address this limitation, we leverage demonstration trajectories generated by the existing centralized planner and introduce a neural network-based surrogate that approximates the mapping from task specifications to cooperative trajectories. 
The surrogate enables rapid generation of decent coordinated trajectories, but, as it is a learned approximation of an optimal solution rather than an exact model-based optimizer, it cannot provide strict guarantees on constraint satisfaction or optimality. To remedy this, we use the surrogate’s predictions solely as an initial solution (learning-based warm start) for the downstream trajectory optimization problem, which then refines them into dynamically feasible and optimal trajectories with significantly reduced computation time. In the following, we refer to this learning-based warm start generator as the surrogate planner.
Figure~\ref{fig:learning_augmented_overview} provides an overview of this learning-augmented planning workflow. 
In summary, this work makes the following novel contributions:
\begin{itemize}
\item This work presents a surrogate planner based on an artificial neural network that maps task specifications to cooperative trajectories as informed warm starts, using separately trained per-agent networks rather than a single monolithic policy to generate both UAV and UGV trajectories.
\item Overall, using these networks, this paper proposes a learning-augmented pipeline for a multi-robot trajectory optimization framework for handover tasks that significantly reduces computation time while preserving feasibility and solution quality.
\end{itemize}

The rest of this paper is structured as follows. Section~\ref{sec:related_work} reviews relevant literature on learning-based approaches. Section~\ref{sec:problem_formulation} formalizes the cooperative handover mission and the underlying model-based trajectory optimization framework used for expert data generation. Section~\ref{sec:methodology} presents the proposed methodology, detailing the relative coordinate transformations, the decoupled LSTM-based surrogate architecture, and the corresponding training paradigm. Section~\ref{sec:numerical_results} evaluates the performance of the learning-augmented pipeline through comprehensive benchmark simulations. Finally, Section~\ref{sec:conclusion} concludes the paper and discusses potential avenues for future research.

\section{Related Work}\label{sec:related_work}

Learning-based methods have been widely explored in robotics to complement or even replace traditional planning and control pipelines. Trained neural networks are used for trajectory prediction in complex environments~\cite{Xu2024}, to directly output collision-free motions that substitute classical motion planners~\cite{Kou2023}, or to implement end-to-end policies that map observations directly to control commands, bypassing explicit trajectory generation and conventional controllers~\cite{Shiyu2025,Sajja2025}. 
Rather than fully discarding well-established planning and control concepts, a complementary line of work employs neural networks as augmented components. Examples include learning-based model predictive control schemes that integrate data-driven models while preserving safety and constraint handling~\cite{Hewing2020} and surrogate trajectory planners that provide informed initializations to accelerate sequential convex programming without compromising solution quality~\cite{Banerjee2025}. 
Since the performance of purely learned policies is often sensitive to distribution shift between training and deployment, as well as modeling errors, directly executing their predictions may lead to unexpected results or even task failure in robotic applications~\cite{Ross2011,Diaz2021}. In addition, learning-based controllers without an explicit control-theoretic scaffold typically lack formal guarantees on stability and constraint satisfaction~\cite{Hewing2020}.
Therefore, this work follows a paradigm, where a learning-based approach is used to augment (in our case accelerate) a model-based approach instead of fully replacing it. The surrogate planner serves as an informed warm start for the optimization-based trajectory planning framework, yielding a fast converged solution without sacrificing feasibility and safety.

A related line of work uses a centralized expert to train decentralized per-agent policies via imitation learning. In such approaches, a centralized controller or planner with access to the global state first generates demonstrations, and each agent then learns a local policy from these expert trajectories~\cite{Lin2022,Agarwal2025}. Subsequently, these learned policies are typically deployed directly, without any further optimal-control-based refinement. In contrast, our design keeps the centralized optimal control problem in the loop, while still benefiting from lightweight per-agent inference. 

\section{Problem Formulation}\label{sec:problem_formulation}

This work addresses a cooperative multi-robot trajectory planning problem for handover tasks, as studied previously in~\cite{Chen2024} without any learning-based process augmentation. Given the initial states $\bm{x}_{\text{uav}}(0)$, $\bm{x}_{\text{ugv}}(0)$ and goal states $\bm{x}_{\text{uav}}(N)$, $\bm{x}_{\text{ugv}}(N)$ of both the quadrotor and the mobile robot, the objective is to generate coordinated trajectories that achieve a successful handover. We focus exclusively on the trajectory generation problem and do not discuss here the detailed grasping process during the handover. Consequently, we assume that the handover task is successfully completed as long as both robots are sufficiently close to one another at a certain timestep.
\subsection{Handover Trajectory Planner}
The state of the quadrotor is defined as $\bm{x}_\text{uav} := \left[\bm{p}_\text{uav}^\top \ \bm{\xi}^\top \ \bm{v}^\top \ \bm{\omega}_\text{B}^\top \right]^\top \in \mathbb{R}^{12} $,
where $\bm{p}_\text{uav}:= \left[x_\text{uav}\ y_\text{uav}\ z_\text{uav}\right]^\top$ denotes the position, $\bm{v}=\dot{\bm{p}}_\text{uav}:=\left[v_x \ v_y \ v_z \right]^\top$ the velocity, $\bm{\xi}:=\left[\phi \ \theta \ \psi\right]^\top$ the Euler angles and $\bm{\omega}_\text{B}$ the vector of angular velocities. Note that we use the notation $(\cdot)_\text{B}$ for quantities expressed in the coordinates of the body-fixed frame.  
The control input of the quadrotor is defined as 
$\bm{u}_\text{uav} = \left[f_{\text{1}} \ f_{\text{2}} \ f_{\text{3}} \ f_{\text{4}}\right]^\top$, where $f_{i}$ is the thrust generated by the $i$th propeller, and the collective thrust along the $z$-axis of the quadrotor's body-fixed frame is given by $f_\text{thrust} = \sum_{n = 1}^{4}f_{i}$. 
The quadrotor dynamics can then be written as
\begin{equation}
  \label{eq:quad_system_dynamics}
  \dot{\bm{x}}_\text{uav} = \begin{bmatrix}
    \bm{v} \\
    \bm{T}(\bm{\xi})\bm{\omega}_\text{B} \\
    \bm{R}_\text{B}(\bm{\xi}) \left[0 \ 0 \ f_\text{thrust}\right]^\top - \left[0 \ 0 \ m_\text{uav}g\right]^\top \\
    \bm{J}_\text{uav}^{-1}(-\bm{\omega}_\text{B}\times\bm{J}_\text{uav}\bm{\omega}_\text{B} + \bm{\tau}_\text{B}(\bm{u}_\text{uav}))
  \end{bmatrix},
\end{equation}
where $m_\text{uav}$ is the quadrotor mass, $\bm{J}_\text{uav}$ is its inertia matrix, $\bm{R}_\text{B}(\bm{\xi})$ is the rotation matrix from the body-fixed frame to the inertial frame, $\bm{\tau}_\text{B}(\bm{u}_\text{uav})$ denotes the control torque acting on the UAV, and $\bm{T}(\bm{\xi})$ maps the body-frame angular velocity to the time derivative of the Euler angles. 
The mobile robot model is based on our omnidirectional mobile platform \textit{HERA}~\cite{Ebel2021}. Its state is defined as $\bm{x}_\text{ugv} := [x_\text{ugv}\ y_\text{ugv}\ v_{x,\text{ugv}}\ v_{y,\text{ugv}}]^\top \in \mathbb{R}^4$, where $x_\text{ugv}, y_\text{ugv}$ are the planar position coordinates and $v_{x,\text{ugv}}, v_{y,\text{ugv}}$ are the corresponding velocity components. Due to the omnidirectional character of the robot, we do not consider its orientation as a state. Its control input is given by $\bm{u}_\text{ugv} := [f_\text{ugv}\ \zeta]^\top \in \mathbb{R}^2$,
where \(f_\text{ugv}\) is the propulsion force in the horizontal plane and \(\zeta\) the angle between this force and the inertial \(x\)-axis. The mobile robot dynamics can be expressed as
\begin{equation}
    \label{eq:omni_dynamics}
    \dot{\bm{x}}_\text{ugv} =
    \begin{bmatrix}
        v_{x,\text{ugv}} \\
        v_{y,\text{ugv}} \\
        f_\text{ugv} \cos(\zeta) / m_\text{ugv} \\
        f_\text{ugv} \sin(\zeta) / m_\text{ugv}
    \end{bmatrix},
\end{equation}
where \(m_\text{ugv}\) denotes the robot mass. Further modeling details can be found in~\cite{Chen2025}.

The primary objective of this work is not to design a new optimization scheme, but to accelerate an existing one. Therefore, we do not reiterate the design details of the optimization problem or its handover-related constraints, and instead directly adopt the discrete-time formulations from~\cite[Eqs.~(10)--(12)]{Chen2024} with the optimization variables $\varepsilon_k, \nu_k \ \text{and} \ \kappa_k$. 

As mentioned, we use in this work the centralized optimization-based trajectory planner proposed in~\cite{Chen2024} to generate a dataset of expert demonstrations.
For the studied UAV-UGV handover task, the optimization problem to be solved reads
\begin{equation}
  \label{eq:final_optimization}
  \begin{aligned}
    \min_{\bm{x}^*} \hspace{3mm} & J_\text{d} \left(\Delta t_k, \bm{x}_{\text{uav},k}, \bm{x}_{\text{ugv},k}, \bm{u}_{\text{uav},k}, \bm{u}_{\text{ugv},k}, \kappa_k, \nu_{k} \right) &
    \\
    \text{s.t.} \hspace{3mm} & \bm{x}_{\text{uav},0} = \bm{x}_{\text{uav}}(0), \hspace{2mm} \bm{x}_{\text{ugv},0} = \bm{x}_{\text{ugv}}(0) &
    \\
    & \bm{x}_{\text{uav},N} = \bm{x}_{\text{uav}}(N), \hspace{2mm} \bm{x}_{\text{ugv},N} = \bm{x}_{\text{ugv}}(N) &
    \\
    & \bm{x}_{\text{uav},k+1} = f_\text{RK}(\bm{x}_{\text{uav},k}, \bm{u}_{\text{uav},k}, \Delta t_k) & \forall  k\in \{0, \ldots, N-1\}, &
    \\
    & \bm{x}_{\text{ugv},k+1} = f_\text{RK}(\bm{x}_{\text{ugv},k}, \bm{u}_{\text{ugv},k}, \Delta t_k) &\forall  k\in \{0, \ldots, N-1\},
    \\
    & \bm{x}_{\text{uav},\text{max}} \leq \bm{x}_{\text{uav},k} \leq \bm{x}_{\text{uav},\text{max}} & \forall  k\in \{0, \ldots, N-1\},
    \\
    & \bm{x}_{\text{ugv},\text{max}} \leq \bm{x}_{\text{ugv},k} \leq \bm{x}_{\text{ugv},\text{max}} & \forall  k\in \{0, \ldots, N-1\},
    \\
    & \bm{u}_{\text{uav},\text{max}} \leq \bm{u}_{\text{uav},k} \leq \bm{u}_{\text{uav},\text{max}} &\forall  k\in \{0, \ldots, N-1\},
    \\
    & \bm{u}_{\text{ugv},\text{max}} \leq \bm{u}_{\text{ugv},k} \leq \bm{u}_{\text{ugv},\text{max}} & \forall  k\in \{0, \ldots, N-1\},
    \\
    & \text{handover task constraints~\cite[Eqs.~(10)--(12)]{Chen2024}},
    \\
    & 0 \leq \varepsilon_{k} \leq 1 &\forall  k\in \{0, \ldots, N-1\}, 
    \\
    & 0 \leq \nu_{k} \leq \nu_\text{max} & \forall  k\in \{0, \ldots, N-1\}, 
    \\
    & \Delta t_\text{min} \leq \Delta t_k \leq \Delta t_\text{max} &\forall  k\in \{0, \ldots, N\}
  \end{aligned}
\end{equation}
with the optimization variable vector defined as
$
  \bm{x}^* = \begin{bmatrix}
    \bm{x}_{0}^*  & \ldots  & \bm{x}_{N}^*
  \end{bmatrix}^\top,
$
where 
\begin{equation*}
  \label{eq:variable_definition}	
  \renewcommand{\arraystretch}{1}
	\bm{x}_k^* = 
    \left\{ 
      \begin{array}{ll}
		    \displaystyle \begin{bmatrix}\bm{x}_{\text{uav},k}^{*\ \top}\ \bm{x}_{\text{ugv},k}^{*\ \top}\ \bm{u}_{\text{uav},k}^{*\ \top}\ \bm{u}_{\text{ugv},k}^{*\ \top}\ \varepsilon_k^*\ \kappa_{k}^*\ \nu_k^*\ \Delta t_k^*\end{bmatrix}^\top 
          & \text{for} \ \ k\in[0,N), \\[2pt]
		    \displaystyle \begin{bmatrix} \bm{x}_{\text{uav},N}^{*\ \top}\ \bm{x}_{\text{ugv},N}^{*\ \top}\ \kappa_{N}^{*}\ \Delta t_N^{*} \end{bmatrix}^\top
          &\text{for}\ \ k=N,
	\end{array} \right.
\end{equation*}
and the second-order Runge-Kutta method $f_\text{RK}$ is utilized to discretize the quadrotor dynamics~\eqref{eq:quad_system_dynamics} and the mobile robot dynamics~\eqref{eq:omni_dynamics}. To improve mission completion efficiency, in terms of both total time and handover progress, while avoiding excessive UAV spinning or tilting, the objective function $J_\text{d}$ is defined as
\begin{equation}
  \label{eq:cost_fct}
  \begin{aligned}
  J_\text{d} =\, 
  & w_1 \sum_{k=0}^{N}\Delta t_k + w_2 \sum_{k=0}^{N}\kappa_k + w_3 \sum_{k=0}^N \left\Vert\bm{x}_{\text{uav},k}\right\Vert_{2}^{2}.
  \end{aligned}
\end{equation}
Since these objective terms have different physical units and scales, the weights are here, without loss of generality, chosen as $w_1=10\,\mathrm{s}^{-1}$, $w_2=1$, and $w_3=1\,\mathrm{m}^{-2}$, which balances their relative influence in the overall optimization.

\subsection{Expert Data Generation}
To generate expert demonstrations, we solve the nonlinear program~\eqref{eq:final_optimization} using IPOPT within the CasADi framework~\cite{Andersson2018}, adopting the model parameters and solver settings in~\cite{Chen2024}. The generated dataset consists of 5000 successfully solved handover trajectories generated by randomly sampling initial and terminal positions. Specifically, horizontal coordinates are sampled within $\mathcal{X}, \mathcal{Y} = [0, 4]\,\text{m}$, while the UAV's vertical position is restricted to $\mathcal{Z} = [0.5, 4]\,\text{m}$. 
Each handover trajectory consists of $N=40$ time steps, providing state sequences  $\{\bm{x}_{\text{uav},k}\}_{k=0}^{N-1}$, $\{\bm{x}_{\text{ugv},k}\}_{k=0}^{N-1}$, and time intervals $\{\Delta t_k\}_{k=0}^{N-1}$ for surrogate training.

\section{Methodology}\label{sec:methodology}
\begin{figure}[tb]
  \centering
    \includegraphics[width=\linewidth]{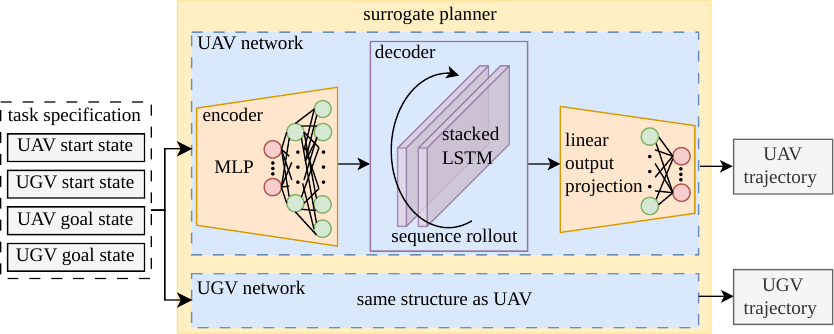}
    \caption{Surrogate architecture with an agent-decoupled encoder–decoder LSTM structure.}
  \label{fig:surrogate_structure}
\end{figure}

For generating robot-specific trajectories, we propose a neural network architecture comprising two structurally identical encoder–decoder LSTM subnetworks, one for the UAV and one for the UGV, see Fig.~\ref{fig:surrogate_structure}. 
Given a global task specification as input, each subnetwork shares the same encoder initialization and recurrently produces a sequence of future predicted states for its corresponding robot. 

\subsection{Expert Data Processing}
To achieve translation invariance and improve learning generalization, we transform the position states in the horizontal plane, i.e., the $x$- and $y$-coordinates, into a localized, UAV-centric relative frame, while keeping the UAV's altitude in the global frame to ensure ground safety. In this way, we avoid representing altitude in a purely relative coordinate, which could otherwise result in generated relative altitudes that drive the UAV below the ground level.
In each expert demonstration, we select the UAV's horizontal start position as a spatial anchor $\bm{a}= [x_{\text{uav},0}\ y_{\text{uav},0}\ 0]^\top \in \mathbb{R}^3$.
At each timestep $k$, the relative state $\bm{s}^{\text{rel}}_k\in \{\bm{x}^{\text{rel}}_{\text{uav},k},\bm{x}^{\text{rel}}_{\text{ugv},k} \}$ is obtained by shifting the corresponding inertial coordinates $\bm{s}^\text{abs}_k\in \{\bm{x}_{\text{uav,k}},\bm{x}_{\text{ugv},k} \}$ according to 
$    \bm{s}^{\text{rel}} = \bm{s}^{\text{abs}} - \bm{C}\bm{a},$
where $\bm{C}$ is a selection matrix that applies the shift only to the horizontal $x$ and $y$ components. Let $\bm{A} \in \mathbb{R}^{4 \times 3}$ be defined by $A_{11} = A_{22} = 1$ and $A_{ij} = 0$ otherwise. For the UGV state, $\bm{C} = \bm{A}$. For the UAV, we set $\bm{C} = [\bm{A}\ \bm{0}_{8\times 3}]^\top$. 
After the spatial shift, the input vector of our surrogate planner is constructed by concatenating the relative start and goal states
$
    \bm{\tau} =: [
    {\bm{x}^{\text{rel}}_{\text{ugv},0}}^\top\
    {\bm{x}^{\text{rel}}_{\text{uav},0}}^\top\
    {\bm{x}^{\text{rel}}_{\text{ugv},N}}^\top\ 
    {\bm{x}^{\text{rel}}_{\text{uav},N}}^\top\
    ]^\top\in \mathbb{R}^{32}
$. Correspondingly, the joint expert trajectory $\mathcal{Y} \in \mathbb{R}^{17\times N}$ is defined as a sequence of state vectors over $N=40$ steps, i.e.,
$\mathcal{Y} = \begin{bmatrix}
\bm{y}_0\ \bm{y}_1\ \dots\ \bm{y}_{N-1}
\end{bmatrix} $, 
$\bm{y}_k = [
    {\bm{x}^{\text{rel}}_{\text{uav},k}}^\top\ 
    {\bm{x}^{\text{rel}}_{\text{ugv},k}}^\top\ 
$
$ \Delta t_k]^\top \in \mathbb{R}^{17} $ at timestep $k$.
To optimize the training performance, we standardize all input features $\bm{\tau}$ and output targets $\mathcal{Y}$ using z-score normalization based on the dataset-wide empirical mean and standard deviation.

\subsection{Surrogate Architecture}

The surrogate architecture is based on an agent-decoupled framework, where the individual subnetworks 
predict robot-specific trajectories for their corresponding UAV or UGV, $\hat{\mathcal{Y}}_{\text{uav}} = \begin{bmatrix}
  \hat{\bm{y}}_{\text{uav},0}\ \hat{\bm{y}}_{\text{uav},1}\ \dots\ \hat{\bm{y}}_{\text{uav},39}
\end{bmatrix} \in \mathbb{R}^{13 \times 40}$ and $\hat{\mathcal{Y}}_{\text{ugv}}= \begin{bmatrix}
  \hat{\bm{y}}_{\text{ugv},0}\ \hat{\bm{y}}_{\text{ugv},1}\ \dots\ \hat{\bm{y}}_{\text{ugv},39}
\end{bmatrix} \in \mathbb{R}^{5 \times 40}$. 
At each timestep $k$, the predicted outputs are given by $\hat{\bm{y}}_{\text{uav},k} = [ \hat{\bm{x}}_{\text{uav},k}^{\text{rel}\,\top}\,$ $\Delta t_k]^\top$ and $\hat{\bm{y}}_{\text{ugv},k} = [ {\hat{\bm{x}}^{\text{rel}\,\top}_{\text{ugv},k}}\, \Delta t_k]^\top$.
For notational consistency, all hatted quantities $\hat{(\cdot)}$ denote predictions by the surrogate planner.
This decoupling allows the networks to specialize in the heterogeneous dynamics of each robot while receiving the same global mission context $\bm{\tau}$. Noting that we use a relative state representation, the anchor point $\bm{a}$ is added to the predicted trajectories to recover their absolute coordinates before they are used as a warm start for the trajectory optimizer. Since these subnetworks are identical in structure, we describe them using a unified notation.

The encoder $E: \mathbb{R}^{32} \to \mathbb{R}^{2 \times L \times d_{\text{hid}}}$  in our framework is implemented as a deep multi-layer perceptron (MLP) to handle the static nature of the task specification. Given the standardized task specification, the encoder initializes the decoder's hidden dynamical states by predicting the optimal initial hidden states $\bm{h}_0 \in \mathbb{R}^{L \times d_{\text{hid}}}$ and the cell states $ \bm{c}_0 \in \mathbb{R}^{L \times d_{\text{hid}}}$ for each robot agent via $\bm{h}_0 = \text{MLP}_h(\bm{\tau})$ and $\bm{c}_0 = \text{MLP}_c(\bm{\tau})$, where the $\text{MLP}_{(\cdot)}$ consists of two linear layers with ReLU activations, projecting the task context into a hidden state dimension of $L \times d_{\text{hid}}$ with $L=2$ layers and $d_{\text{hid}}=256$ hidden units.
Then, these initial hidden states $(\bm{h}_0, \bm{c}_0)$ are passed as the starting configuration to a stacked LSTM-based decoder $D: \mathbb{R}^{d_{\text{out}}} \times \mathbb{R}^{2 \times L \times d_{\text{hid}}} \to \mathbb{R}^{d_{\text{out}}} \times \mathbb{R}^{2 \times L \times d_{\text{hid}}} $ for the recurrent trajectory generation, where $d_{\text{out}}$ is set to 13 for the UAV and 5 for the UGV.
At each discrete timestep $k \in \{0, \dots, 39\}$, the recurrent state update is given by $(\bm{h}_{k}, \bm{c}_{k}) = D(\mathbf{in}_k, \bm{h}_{k-1}, \bm{c}_{k-1})$ and the subsequent output projection is computed as $\hat{\bm{y}}_k = \bm{W}_{\text{out}} \bm{h}_{k} + \bm{b}_{\text{out}}$, where $(\bm{h}_k, \bm{c}_k)$ denotes the recurrent hidden and cell states evolving over the prediction horizon. The output projection is parameterized by weights $\bm{W}_{\text{out}} \in \mathbb{R}^{d_{\text{out}} \times d_{\text{hid}}}$ and biases $\bm{b}_{\text{out}} \in \mathbb{R}^{d_{\text{out}}}$. The term $\mathbf{in}_k$ denotes the input token at timestep $k$, representing either the ground-truth state $\bm{y}_{k-1}$ at the previous timestep from the expert data or the subnetwork's own previous prediction $\hat{\bm{y}}_{k-1}$. 

\subsection{Training Paradigms}
The training procedure follows Algorithm~\ref{alg:training_decoupled}.
\begin{algorithm}[btp]
\caption{Agent-decoupled Training}
\label{alg:training_decoupled}
\begin{algorithmic}[1]
\STATE \textbf{Initialize:} encoders $E_{U}$ and decoders $D_{U}$ for all $U \in \{\mathrm{UAV}, \mathrm{UGV}\}$, $K_{\text{max}}$, empirical mean and standard deviation $(\bm{\mu}, \bm{\sigma})$ on the training set
\FOR{epoch $K = 1$ to $K_\text{max}$}
    \STATE compute $P_{\text{tf}}$ based on Eq.~\eqref{eq:scheduled_sampling}
    \FOR{each batch $(\bm{\tau}, \mathcal{Y}, g)$}
        \STATE standardize the input vector $\bm{\tau}$ and expert trajectories $\mathcal{Y}$ using $(\bm{\mu}, \bm{\sigma})$
        \FOR{each robot agent $U \in \{\text{UAV, UGV}\}$}
            \STATE initialize $(\bm{h}_{U,0}, \bm{c}_{U,0}) = E_{U}(\bm{\tau})$ 
                    and set $\mathbf{in}_{U,0} \leftarrow \text{start state from } \bm{\tau}$
            \FOR{step $k = 1$ to $39$}
                \STATE predict $\hat{\bm{y}}_{U,k}$, and update $(\bm{h}_{U,k}, \bm{c}_{U,k}) = D_{U}(\mathbf{in}_{U,k}, \bm{h}_{U,k-1}, \bm{c}_{U,k-1})$
                \STATE sample $r \sim \text{uniform}(0,1)$ and set
                \[
                    \mathbf{in}_{U,k+1} =
                    \begin{cases}
                        \bm{y}_{U,k}, & r < P_{\text{tf}} \\
                        \hat{\bm{y}}_{U,k}, & \text{otherwise}
                    \end{cases}
                \]
            \ENDFOR
            \STATE compute $\mathcal{L}_U$ using Eq.~\eqref{eq:weighted_mse_loss}$\leftarrow \text{WeightedMSE}(\hat{\mathcal{Y}}_{\text{U}}, \mathcal{Y}_U, g)$ 
            and update $E_{U}$, $D_{U}$ 
        \ENDFOR
    \ENDFOR
    \STATE \textbf{Validation:}
    evaluate in pure autoregressive mode ($P_{\text{tf}}=0$) and adjust learning rate via \textit{ReduceLROnPlateau} based on validation loss
\ENDFOR
\end{algorithmic}
\end{algorithm}
The model minimizes a mean squared error (MSE) loss with time-dependent weights in the form
\begin{equation}
  \label{eq:weighted_mse_loss}
  \mathcal{L} = \frac{1}{BN} \sum_{i=1}^{B} \sum_{k=0}^{N-1} w_k \| \bm{y}_{i,k} - \hat{\bm{y}}_{i,k} \|_2^2 \,,
\end{equation} 
where $B=256$ is the batch size and $N=40$ the horizon. We use the weights to prioritize critical phases during the handover mission, $w_0=5$ for the initial state, $w_{N-1}=10$ for the goal state, and $w_{[g-2, g+2]}=35$ within a time window $[g-2, g+2]$ around the expert handover time index $g$, and $w_k=1$ otherwise.

To bridge the gap between training and inference,
scheduled sampling is employed. The probability of using teacher forcing $P_{\text{tf}}$ decays as described by
\begin{equation}
  \label{eq:scheduled_sampling}
    P_{\text{tf}} = \max \left( 0.1, 1.0 - \frac{K}{0.75 \times K_{\text{max}}} \right),
\end{equation}
where $K$ is the current epoch and $K_{\text{max}}=200$ is the total number of epochs.
The decoupled UAV and UGV subnetworks are trained separately using the Adam optimizer, with an initial learning rate of $10^{-3}$ that is adaptively adjusted by a scheduler based on validation performance.

\section{Numerical Results}\label{sec:numerical_results}
This section evaluates the performance of the proposed learning-augmented trajectory optimization framework. 
For a qualitative assessment, we summarize the results of trajectory optimization~\eqref{eq:final_optimization} with and without learning-based warm starts over 100 randomized benchmark runs in Table~\ref{tab:benchmark_results}.
\begin{table}[tbp]
\centering
\caption{Statistical comparison with and without learning-based warm starts.}
\label{tab:benchmark_results}
\begin{tabular}{@{}l@{\hspace{5mm}}l@{\hspace{5mm}}l@{}}
\toprule
metric & without warm start & with warm start \\
\midrule
average iterations & 230.01 & 146.33 \\
minimum iterations & 61 & 73 \\
maximum iterations & 1409 & 855 \\
\midrule
average time cost ($\Delta t_k$)  & 279.94 & 273.64 \\
average kappa cost ($\kappa_k$) & 18.01 & 19.22 \\
average state cost ($\left\Vert\bm{x}_{\text{uav},k}\right\Vert$) & 283.05 & 290.28 \\
\midrule
success rate & 96.0\% & 100.0\% \\
average computation time & 4.3815\,[s] & 1.4436\,[s] \\
\bottomrule
\end{tabular}
\end{table}
\begin{figure}[tb]
  \centering
  \includegraphics[width=0.8\linewidth]{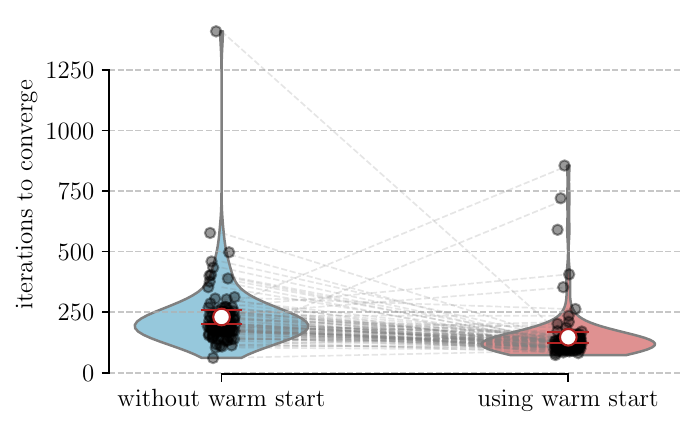}
  \caption{Comparison of convergence iterations for cold and warm starts. Each pair of connected points represents a matched run, i.e., a cold and warm start for the same start and goal states. The white circular markers indicate empirical means, with red bars showing 95\% confidence intervals.}
  \label{fig:iteration_comparison}
\end{figure}
The learning-based warm start framework reduces the average number of optimization-algorithm iterations by approximately 36\% and yields a 3.04$\times$ speedup in computation time from 4.38\,s to 1.44\,s, while maintaining comparable solution quality, as indicated by the nearly identical cost components.
A raincloud plot in Fig.~\ref{fig:iteration_comparison} visualizes the iterations to convergence, where the colored violin-shaped densities depict the distribution of iterations for runs with and without warm starts. We plot the iterations required for 96 successful runs without warm starts and 100 successful runs with warm starts as individual points and connect matched trials with dashed lines. Because only successful optimizations are shown, four warm-started trials do not have corresponding baseline solutions and remain unmatched.
As the highly nonlinear and nonconvex centralized trajectory optimization is sensitive to the initial guess, the benefits of providing reliable warm starts are clearly reflected in the reduced iteration counts. Although a few outliers near the boundary of the training distribution require more iterations than their cold-start counterparts, as reduced surrogate accuracy may place the initial guess less optimally, the otherwise quite consistent decrease across different handover missions with randomized start and goal states indicates that the neural surrogate planner provides a universally superior initial guess for the optimizer.

\begin{figure}[bt]
  \centering
  \includegraphics[width=\linewidth]{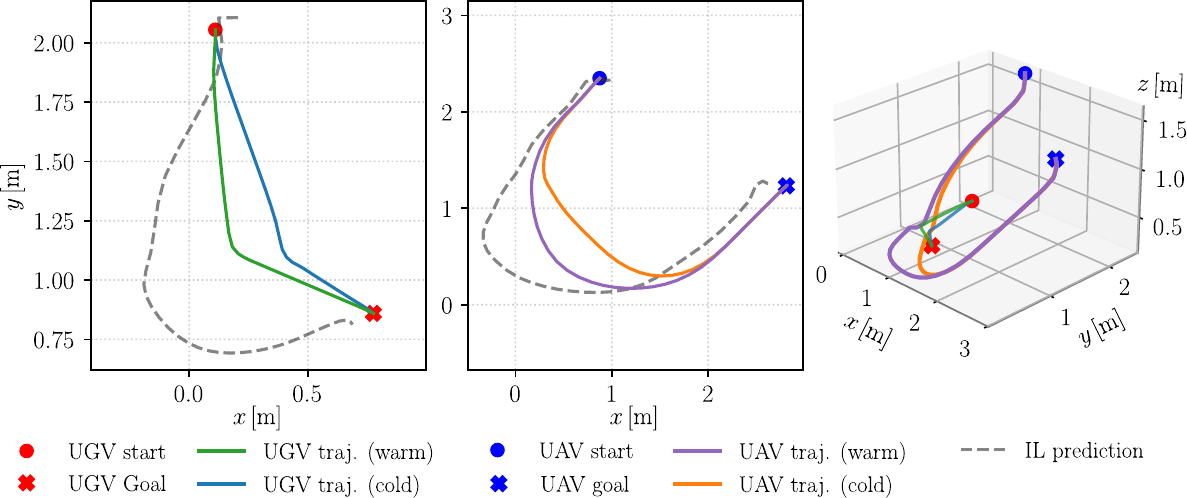}
  \caption{Representative coordinated trajectories for a randomly selected handover mission, illustrating the raw predictions from the surrogate planner (IL prediction), the refined trajectories obtained by using these predictions as a warm start (warm), and the solutions from a cold-start optimization (cold).}
  \label{fig:trajectory_strategy_comparison}
\end{figure}

To provide a direct and intuitive illustration, we visualize the trajectories for a randomly selected handover mission in Fig.~\ref{fig:trajectory_strategy_comparison}. 
In our approach, the surrogate planner produces a coordinated initial guess, shown as gray dashed lines, in which the UAV and UGV already exhibit the required synergy for the handover. This informed initialization places the subsequent model-based optimizer in a favorable region of attraction, leading to faster convergence and fewer failures compared to starting from a naive initialization.
Quantitatively, the learning-based warm start increases the success rate from 96.0\% to 100.0\%. This improvement indicates that the surrogate-initialized trajectories help the solver avoid non-convergence and reduce the likelihood of violating problem constraints.

\section{Conclusion}\label{sec:conclusion}

In this work, we introduce a learning-augmented pipeline to accelerate centralized trajectory optimization for heterogeneous robotic handover tasks. By leveraging expert data generated from an existing centralized optimization-based planner to train the proposed decoupled LSTM-based surrogate subnetworks, the generated coordinated initial guesses effectively capture the synergy between UAV and UGV. Using these initial guesses, our approach achieves a more than 60\% reduction in computational overhead and improves the solver success rate to 100\% compared to a standard cold start procedure, all while preserving solution quality. As far as we are aware, this is the first work that combines agent-wise learned surrogates with a centralized trajectory optimization framework in a cooperative setting to generate dynamically feasible and optimally coordinated UAV-UGV handover trajectories.
Future research will extend this framework to complex environments with dynamic obstacles and include real-world experimental validation.

{\small\medskip\noindent\textbf{Acknowledgements} This research was supported by the German Research Foundation under Grants 433183605 and 501890093, and through Germany’s Excellence Strategy under Grant EXC 2075-390740016.}

%

\end{document}